\documentclass[11pt]{article}

\usepackage[final]{acl}

\usepackage{times}
\usepackage{latexsym}

\usepackage[T1]{fontenc}
\usepackage{amssymb}

\usepackage[utf8]{inputenc}

\usepackage{microtype}

\usepackage{inconsolata}

\usepackage{graphicx}
\usepackage{hyperref}
\usepackage{url}
\usepackage{multirow}
\usepackage[table]{xcolor}
\usepackage{booktabs}
\usepackage{graphicx}
\usepackage{amsmath}
\usepackage{placeins}
\title{SCULPT: Constraint-Guided Pruned MCTS that Carves Efficient Paths for Mathematical Reasoning}

\author{
  Qitong Fang\thanks{Equal contribution and corresponding author.} \\
  Jilin Jianzhu University \\
  \texttt{fangqitong@}\\\texttt{student.jlju.edu.cn}
  \\\And
  Haotian Li\thanks{Equal contribution and corresponding author.} \\
  Jilin Jianzhu University \\
  \texttt{lihaotian@}\\\texttt{student.jlju.edu.cn}
  \\\And
  Xu Wang\thanks{Corresponding author.} \\
  Jilin Jianzhu University \\
  \texttt{wangxu@}\\\texttt{jlju.edu.cn}
}

\begin{document}
\maketitle
\begin{abstract}
Automated agent workflows can enhance the problem-solving ability of large language models (LLMs), but common search strategies rely on stochastic exploration and often traverse implausible branches. This occurs because current pipelines sample candidate steps from generic prompts or learned policies with weak domain priors, yielding near-random walks over operators, units, and formats. To promote ordered exploration, this paper introduces SCULPT, a constraint-guided approach for Monte Carlo Tree Search (MCTS) that integrates domain-aware scoring into selection, expansion, simulation, and backpropagation. SCULPT scores and prunes actions using a combination of symbolic checks (dimensional consistency, type compatibility, magnitude sanity, depth control, and diversity) and structural pattern guidance, thereby steering the search toward plausible reasoning paths. Under matched LLM configurations, SCULPT yields stable improvements on multiple datasets; additional results with GPT-5.2 assess executor transferability and performance on frontier reasoning models. Overall, domain-aware constraints can improve accuracy while maintaining efficiency and reasoning stability.
\end{abstract}

\section{Introduction}

The rapid progress of LLMs has enabled automated agent workflows to structure complex problem solving~\citep{yu2025survey,tan2025meta,ferrag2025llm,li2024survey,barbosa2024collaborative}. However, existing systems predominantly rely on stochastic exploration~\citep{hao2023exploration,peng2023towards}, where candidate steps are sampled from generic prompts or weakly informed policies lacking domain-specific priors~\citep{tarbouriech2021provably}. This unguided sampling often leads to computational waste on mathematically implausible branches, such as those violating type or unit consistency, thereby hindering search convergence~\citep{karras2024guiding}.

Mathematical reasoning poses unique challenges due to its heterogeneous structure across domains like arithmetic, geometry, and number theory~\citep{ahn2024large,wang2024measuring}. Reasoning paths must adhere to strict typing rules, unit conventions, and algebraic identities. Traditional stochastic search suffers from excessive branching factors where many actions are admissible but logically unsound~\citep{li2023reinforcement}. Moreover, long-horizon dependencies and sensitivity to formatting conventions~\citep{karpukhin2024hotpp,shao2025symskill} frequently result in noisy rollout estimates and avoidable errors, even when utilizing high-capacity executors~\citep{chen2024chronos}.

To address these limitations, we propose a paradigm of constraint-guided exploration. By integrating domain-aware scoring functions into the search process, we provide local structure to the search space, prioritizing logically consistent steps while preserving necessary diversity. This approach transforms unguided sampling into an ordered exploration of plausible operations. Accordingly, we introduce SCULPT, a framework that integrates dimensional, type, pattern, and magnitude constraints throughout the workflow optimization loop. SCULPT employs compliance-aware selection, thresholded expansion, and reward shaping to steer the search towards high-probability reasoning trajectories.


\begin{figure*}[!t]
\centering
\includegraphics[width=1\linewidth]{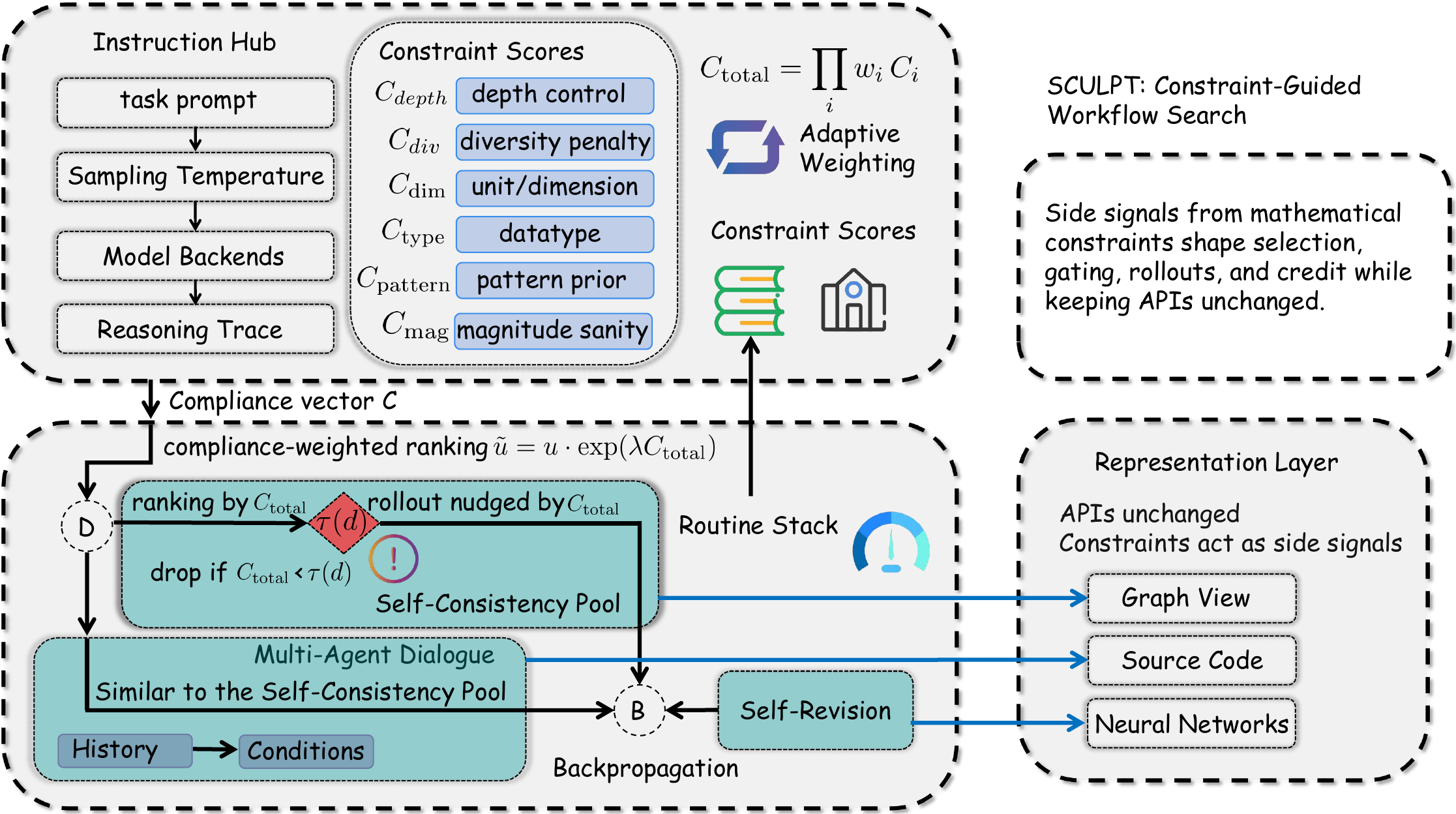}
\caption{SCULPT architectural overview. Upper left: The \textbf{Instruction Hub} decomposes problem requirements into a vector of symbolic constraints (dimensional consistency $C_{\mathcal{U}}$, type compatibility $C_{\mathcal{T}}$, pattern similarity $C_{\mathcal{P}}$, magnitude sanity $C_{\mathcal{M}}$, depth control $C_{\mathcal{D}}$, and diversity $C_{\mathcal{V}}$). Lower left: The \textbf{Routine Stack} executes the MCTS search loop (selection, expansion, simulation, backpropagation) shaped by the aggregate compliance \(C_{\text{total}}\). Lower right: Optimized programs are exposed via the \textbf{Representation Layer}, which provides multiple views including symbolic logic graphs and computational flow visualizations.}
\label{fig:node_operator_edge_constrained}
\end{figure*}

\begin{figure}[!t]
\centering
\includegraphics[width=1\linewidth]{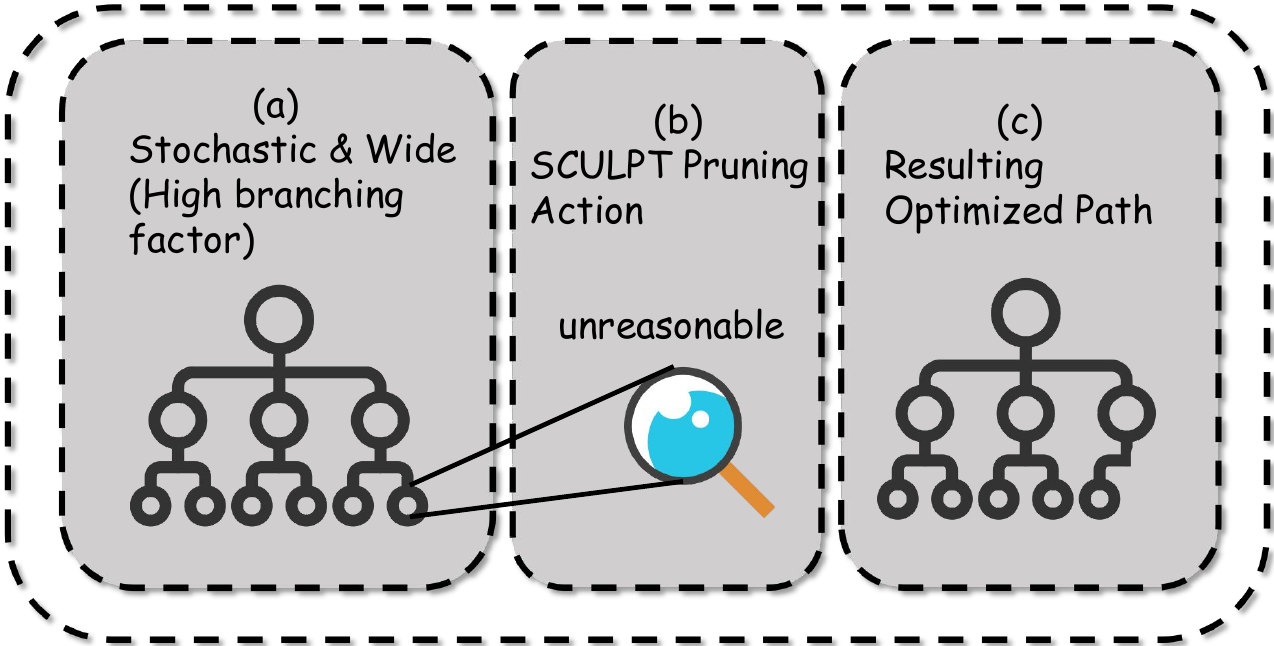}
\caption{Search space transformation via SCULPT. (a) A vanilla MCTS-based workflow search can suffer from a high branching factor due to many admissible but mathematically implausible actions. (b) SCULPT identifies and prunes these nodes early through symbolic constraint checks. (c) The resulting search space is regularized, focusing computational budget on logically consistent reasoning paths.}
\label{fig:search_space_transformation}
\end{figure}

The contributions of this work are three-fold: (1) We formalize mathematical workflow search as a constraint-guided optimization problem, integrating domain-aware scoring into the MCTS loop. (2) We introduce SCULPT, which incorporates six constraint families (combining static symbolic rules and structural pattern guidance) across the selection, expansion, simulation, and backpropagation phases. (3) We present an adaptive weighting mechanism that dynamically prioritizes constraints based on their correlation with validation performance. Empirically, SCULPT yields systematic improvements in accuracy across GSM8K, MATH, and GSM-Hard while reducing optimization overhead by pruning 34.2\% of implausible branches.

\section{Related Work}
\label{gen_inst}

Recent work on automating agent workflows spans prompt-level optimization \citep{fernando2023promptbreeder,yuksekgonul2024textgrad,tang2023verifai,khattabsri}, system/hyperparameter optimization \citep{saad2024archon}, and end-to-end workflow optimization \citep{li2024autoflow,zhou2024symbolic,zhuge2024gptswarm,hu2024automated}. MCTS-based workflow optimizers, such as AFlow \citep{zhang2024aflow}, scale search by representing complete workflows as tree nodes. SCULPT employs the canonical four-stage MCTS loop (selection, expansion, simulation, and backpropagation) and introduces constraint-guided exploration by integrating compliance signals derived from symbolic constraints to modulate rewards and prune branches throughout the search process.

Formally, given a dataset of problems and an initial workflow, the objective is to synthesize a program that maximizes accuracy under a fixed optimizer--executor budget. SCULPT operationalizes this by computing constraint compliance scores that regularize the search. As illustrated in Figure~\ref{fig:node_operator_edge_constrained}, the Instruction Hub decomposes requirements into a compliance vector, while the Routine Stack executes a shaped search loop. This architecture allows constraints to act as side signals that guide exploration without altering the underlying executor APIs.

The implementation of SCULPT follows three core principles: \emph{early hard pruning}, \emph{soft shaping}, and \emph{execution-aware constraints}. During expansion, depth-aware thresholds filter candidates with low compliance before they enter the tree, systematically narrowing the search space (Figure~\ref{fig:search_space_transformation}). Simultaneously, compliance scores modulate selection and simulation to bias the search toward structurally sound steps. Finally, magnitude constraints are applied during simulation to ensure numerical sanity. Through these mechanisms, we examine whether domain-aware constraint integration can improve reasoning stability and efficiency across benchmarks such as GSM8K and MATH, particularly in handling complex conventions like ring areas or coordinate ratios.

\section{Methodology}
\label{headings}

\subsection{Problem Formulation and Constraint Aggregation}
Given a dataset of problems and an initial workflow, the goal is to find a program (workflow instance) that maximizes accuracy under a fixed optimizer--executor budget. Let $C_i$ denote constraint scores for each constraint family. SCULPT computes a normalized aggregate compliance $C_{\text{total}}$ using a weighted geometric mean:
\begin{equation}
C_{\text{total}} = \exp\left( \frac{\sum_{i \in \{\mathcal{U}, \mathcal{T}, \mathcal{P}, \mathcal{M}, \mathcal{D}, \mathcal{V}\}} w_i \ln(C_i + \epsilon)}{\sum_{j \in \{\mathcal{U}, \mathcal{T}, \mathcal{P}, \mathcal{M}, \mathcal{D}, \mathcal{V}\}} w_j} \right)
\label{eq:compliance_agg}
\end{equation}
where $\epsilon = 0.01$ is a smoothing constant preventing over-pruning. The denominator ensures proper normalization during adaptive weight updates. This aggregate is used to (a) prune expansions via a depth-aware threshold $\tau(d)$ defined as:
\begin{equation}
\tau(d) = \max(\tau_{\min}, \tau_0 - k \cdot d)
\label{eq:tau_threshold}
\end{equation}
where $\tau_0$, $\tau_{\min}$ and $k$ are hyperparameters controlling the initial gate, floor, and decay rate respectively (we set $\tau_0 = 0.6$, $\tau_{\min} = 0.3$, and $k = 0.05$ in our experiments). This aggregate is also used to (b) modulate the selection score $\tilde{u}$. We define the compliant selection score as:
\begin{equation}
\tilde{u} = u \cdot \exp(\lambda C_{\text{total}}), \quad u = Q + c \cdot U
\label{eq:selection_score}
\end{equation}
where $u$ is the standard UCT score consisting of the value estimate $Q$ and the exploration bonus $U$ weighted by $c$ (we use $c = 1.414$, the standard UCT exploration constant). The parameter $\lambda$ controls the strength of compliance shaping (we set $\lambda = 0.5$ in our experiments). By using a multiplicative exponential shaping, SCULPT biases selection toward structurally consistent workflows while preserving the ordering induced by the base UCT statistics.

\subsection{Architectural Overview}
SCULPT is a domain-aware Monte Carlo Tree Search (MCTS) framework designed to optimize agentic workflows for mathematical reasoning. The system consists of three primary layers: (1) the \textbf{Instruction Hub}, which extracts symbolic constraints (e.g., dimensional checks, type rules) from problem statements; (2) the \textbf{Routine Stack}, where the core search loop (selection, expansion, simulation, and backpropagation) operates over workflow programs; and (3) the \textbf{Representation Layer}, which translates optimized workflows into executable formats and provides visualizations of the reasoning logic.

In MCTS-based workflow optimization, an agentic workflow is typically represented as an executable code artifact (e.g., a structured Python program) that orchestrates LLM calls and deterministic routines. Each node in the MCTS tree corresponds to a \emph{complete workflow program}, enabling search to reason over global structure rather than individual steps. For constraint scoring, SCULPT consumes a lightweight workflow state derived from this code representation (e.g., depth, operator multiset/histogram, and optional schema-derived tags) and, when available, runtime traces (e.g., intermediate numeric magnitudes). An optimization round follows the standard MCTS cycle: selection, expansion (minimal code edits proposed by an optimizer LLM), simulation (executing the candidate workflow on a validation set), and backpropagation. Constraint signals act as side information that shapes this cycle without changing the executor interface.

\subsection{Mathematical Domain Constraints}
SCULPT employs six constraint families to guide the search loop, organized into three conceptual categories: structural complexity, mathematical consistency, and pattern-magnitude sanity. These constraints consist of \emph{static symbolic rules} (dimensional consistency, type compatibility, magnitude sanity, depth control, and diversity) that encode mathematical domain knowledge without requiring training data, and \emph{structural pattern guidance} that captures effective workflow motifs within the optimization process.

\paragraph{Structural Complexity (\(\mathcal{D}, \mathcal{V}\)).} 
To prevent overly redundant workflows, we penalize extreme tree depths and reward operator diversity. The depth penalty $C_{\mathcal{D}}(w)$ penalizes workflows exceeding a maximum depth threshold $d_{\max}=15$ with decay strength $\beta=0.1$:
\begin{equation}
C_{\mathcal{D}}(w) = \max\left(0, 1.0 - \beta \max(0, d(w) - d_{\max})\right)
\label{eq:depth_penalty}
\end{equation}
where $d(w)$ is the maximum depth. The diversity reward $C_{\mathcal{V}}(w)$ measures the Shannon entropy of operator usage to encourage diverse mathematical operations:
\begin{equation}
C_{\mathcal{V}}(w) = \frac{H(\mathbf{p}_w)}{\ln(|O|)}
\label{eq:diversity_reward}
\end{equation}
where $\mathbf{p}_w$ is the normalized frequency vector of operators, $H(\mathbf{p}_w) = -\sum_{i} p_{w,i} \ln(p_{w,i})$, and $|O|$ is the total number of distinct operators. This score ranges from 0 to 1, rewarding balanced operator distributions.

\paragraph{Mathematical Consistency (\(\mathcal{U}, \mathcal{T}\)).} 
Reasoning must maintain internal validity. We enforce \emph{Physical Unit Analysis} (\(\mathcal{U}\)) as a symbolic check $C_{\mathcal{U}}(w)$ by categorizing variables into physical dimensions (e.g., length, time, mass). Operations are checked for compatibility: addition requires matching units, while multiplication and division transform units dimensionally. Calculus operations are modeled as derivative/integral transformations (e.g., differentiating with respect to a variable transforms its dimension accordingly). The compliance score is the ratio of unit-consistent operations:
\begin{equation}
C_{\mathcal{U}}(w)=
\begin{cases}
\begin{aligned}
\tfrac{1}{|O_w^{\mathcal{U}}|}\!\sum_{o\in O_w^{\mathcal{U}}}\!\mathbf{1}\{\mathrm{units\mbox{-}ok}(o)\},\\[-2pt]
\qquad |O_w^{\mathcal{U}}|>0
\end{aligned}\\
0.5,\qquad |O_w^{\mathcal{U}}|=0
\end{cases}
\label{eq:unit_check}
\end{equation}
where $O_w^{\mathcal{U}}$ denotes the subset of operations for which unit-like tags are available from the workflow representation (otherwise this component defaults to a neutral prior). \emph{Structural Type Compatibility} (\(\mathcal{T}\)) verifies data types and tensor ranks when available, enforcing rules such as non-negative inputs for square roots and positive inputs for logarithms:
\begin{equation}
\begin{split}
C_{\mathcal{T}}(w) = \frac{1}{|O_w|} \sum_{o \in O_w} \mathbf{1}\{&\mathrm{shape\mbox{-}ok}(o) \\
&\wedge \mathrm{type\mbox{-}ok}(o)\},
\end{split}
\label{eq:type_check}
\end{equation}
where $O_w$ is the set of operations, and $\mathrm{shape\mbox{-}ok}$ validates linear algebra dimensions (scalar, vector, matrix) for operator inputs.

\paragraph{Pattern and Magnitude Sanity (\(\mathcal{P}, \mathcal{M}\)).} 
The pattern guidance component \(C_{\mathcal{P}}(w, p) = \mathrm{sim}\!\bigl(\mathbf{v}_w, \mathbf{v}_{P}\bigr)\) uses cosine similarity to compare a workflow's operator histogram $\mathbf{v}_w$ against a motif library $\mathbf{v}_{P}$ maintained as part of the optimizer state. The library is initialized with baseline structural templates (10--15 per category) encoding common mathematical reasoning patterns. During optimization, the library is refined every 3 rounds by clustering workflow operator histograms via k-means ($k=20$ per category) and retaining representative motifs that differ by at least 0.3 cosine distance. The library typically contains 80--120 motifs across four MATH categories at optimization convergence. Workflows without matching patterns receive a neutral score of 0.5 during early rounds. Following standard workflow optimization protocol, the motif library is constructed using the validation split during the optimization phase and remains fixed during test evaluation. Additionally, the magnitude of intermediate numerical results $\mathcal{X}_w$ is regularized to suppress pathological scaling. Unlike the static constraints above, magnitude checking requires execution traces and is applied during simulation and backpropagation:
\begin{equation}
\begin{split}
C_{\mathcal{M}}(\mathcal{X}_w) = &\max\left(0, 1.0 - \delta \frac{\max(|\mathcal{X}_w|) - \theta}{\theta}\right) \\
&\quad \cdot \mathbf{1}\{\max(|\mathcal{X}_w|) > \theta\} \\
&\quad + \mathbf{1}\{\max(|\mathcal{X}_w|) \leq \theta\}
\end{split}
\label{eq:magnitude_sanity}
\end{equation}
where $\theta = \max(|\mathbf{V}_{in}|) \cdot 10^{\gamma}$ is dynamically computed from problem constants $\mathbf{V}_{in}$ with scaling margin $\gamma=2$, and $\delta = 0.5$ controls penalty strength.

\subsection{Adaptive Weighting Rule}
To reflect varying constraint importance across tasks, SCULPT employs adaptive weighting. Weights $w_i$ are initialized uniformly as $w_i^{(0)} = 1/6$ for all six constraint families, then updated based on running correlation between constraint scores $\{C_{i,k}\}$ and validation rewards $\{\mathcal{A}_k\}$ stored in buffer $\mathcal{B}$. This allows the system to prioritize constraints that are most discriminative for a given problem set. After a warm-up period of 5 rounds with fixed weights, weights are updated using:
\begin{equation}
\begin{split}
w_i^{(t+1)} \leftarrow &(1-\alpha) \cdot \frac{w_i^{(t)} \exp(\eta \cdot \mathrm{corr}(\mathbf{c}_i, \mathbf{a}))}{\sum_j w_j^{(t)} \exp(\eta \cdot \mathrm{corr}(\mathbf{c}_j, \mathbf{a}))} \\
&\quad + \alpha \cdot w_i^{(0)}
\end{split}
\label{eq:adaptive_weighting}
\end{equation}
where $\mathbf{c}_i$ and $\mathbf{a}$ are vectors of constraint scores and accuracy outcomes, $\eta = 0.1$ controls the adaptation rate, $\alpha = 0.01$ is the decay strength toward initial weights, and correlation is calculated using Pearson correlation over a sliding window of the last $K=10$ evaluated workflows. The decay term maintains exploration stability by preventing weights from deviating too far from their initial uniform distribution.

\section{Implementation and Experimental Setup}
\label{sec:design_details}

SCULPT integrates as a lightweight guidance signal. During each optimization round, the optimizer proposes workflow candidates. In selection, SCULPT computes the compliance vector \(\mathbf{C}\) for static constraints and its aggregate \(C_{\text{total}}\). Candidates below the depth-aware threshold $\tau(d)$ are filtered out; remaining candidates are prioritized by the compliance-shaped score $\tilde{u}$. During simulation, workflows are executed and magnitude constraints evaluated. Backpropagation multiplies simulation returns by the updated \(C_{\text{total}}\). Constraint weights are updated using the correlation-based rule. We fix the random seed to \textbf{42} and use matched optimizer--executor settings for reproducibility.

\subsection{Datasets}
In our experiments, we utilize three publicly available mathematical datasets: GSM8K \citep{cobbe2021training}, MATH \citep{hendrycks2021measuring}, and GSM-Hard \citep{gao2023pal} as our benchmark datasets. To ensure fair comparison with prior workflow optimization research \citet{saad2024archon}, we split our benchmark datasets into validation and test sets in a 1:4 ratio. For the MATH~\citep{hendrycks2021measuring} dataset, consistent with the protocol in~\citet{hong2025data}, we select problems of difficulty level 5 from four typical categories (combinatorics and probability, number theory, pre-algebra, and pre-calculus).

\subsection{Baselines and Metrics}
The reasoning workflows generated by SCULPT are compared against the following established baselines across all experimental cohorts: (1) \textbf{IO}: direct model invocation without any reasoning framework; (2) \textbf{Self-Refine} \citep{madaan2023self}: iterative self-correction where the model critiques and updates its own answers; (3) \textbf{MultiPersona} \citep{wang2024unleashing}: a collaborative reasoning approach where multiple model identities debate to reach a consensus; (4) \textbf{ADAS} \citep{hu2024automated}: an automated framework that searches for agentic reasoning structures; and (5) \textbf{AFlow} \citep{zhang2024aflow}: a strong MCTS-based workflow optimizer. For GSM8K, MATH, and GSM-Hard (v2 split), accuracy (\%) is used as the primary evaluation metric.

\subsection{Experimental Setup}
Three experimental cohorts are evaluated to disentangle the effect of executors and to enable consistent comparisons.

\textbf{Cohort A (GPT-4o-mini executor).} For fair comparison, we adopt a widely used configuration with Claude-3.5-sonnet as the optimizer and GPT-4o-mini-0718 as the executor. Datasets include GSM8K and MATH. To maintain experimental tractability while ensuring comprehensive evaluation, we focus on these two datasets and do not evaluate GSM-Hard in this cohort. For SCULPT, we set the number of optimization rounds to 15 because validation performance typically plateaus by this point in preliminary runs, and additional rounds yield diminishing returns; we therefore use 15 rounds as a near-saturated setting for SCULPT. In contrast, we report results for AFlow and ADAS using their standard configurations (20 and 30 rounds, respectively).

\textbf{Cohort B (GPT-5-mini executor).} To assess transfer and stronger executors, we adopt GPT-5-mini both as the optimizer and the executor for SCULPT. All compared methods use GPT-5-mini as the executor under matched LLM configurations. Results are obtained by running the same experimental protocol with GPT-5-mini, using identical data splits, random seeds, and evaluation procedures as SCULPT. Data splits are aligned with Cohort A unless otherwise noted for GSM-Hard (v2).

\paragraph{Rounds, tokens, and cost accounting.}
We intentionally run each baseline with the optimization-round setting commonly used in its original paper or public implementation, rather than forcing an identical number of rounds across methods. In this setting, achieving higher accuracy with fewer rounds is a practical advantage, reflecting improved sample efficiency of the search process.
To quantify efficiency, we log per-request token usage for \emph{both} the optimizer and executor (prompt/input tokens and completion/output tokens) during the entire pipeline, and report \textbf{Tokens} as the average total tokens consumed \emph{per problem} (end-to-end, across all rounds). We compute \textbf{Cost (\$)} by applying the corresponding model-specific pricing to the logged input/output tokens and averaging per problem; IO reduces to a single executor call under the same logging rule.

\subsection{Ablation Studies}
To address where constraints matter most, we design ablations along two axes. First, we isolate each constraint family by enabling one component at a time while keeping the others neutral, producing depth-only, diversity-only, dimensional-only, type-only, pattern-only, and magnitude-only variants (using GPT-5-mini as the executor). Second, we isolate each injection stage by enabling constraint shaping in only one part of the search loop (using GPT-5.2 as the executor to assess whether the relative importance of different MCTS phases generalizes across model capabilities). In addition, we compare adaptive weighting against fixed weights (using GPT-5-mini). Across variants within each ablation axis, we keep the data split, evaluation protocol, and iteration budget matched, and report accuracy and average token usage per problem for efficiency comparison.

\section{Experiments}
\label{sec:experiments}

\subsection{Ablation Analysis}
\begin{figure*}[!t]
\centering
\includegraphics[width=1\linewidth]{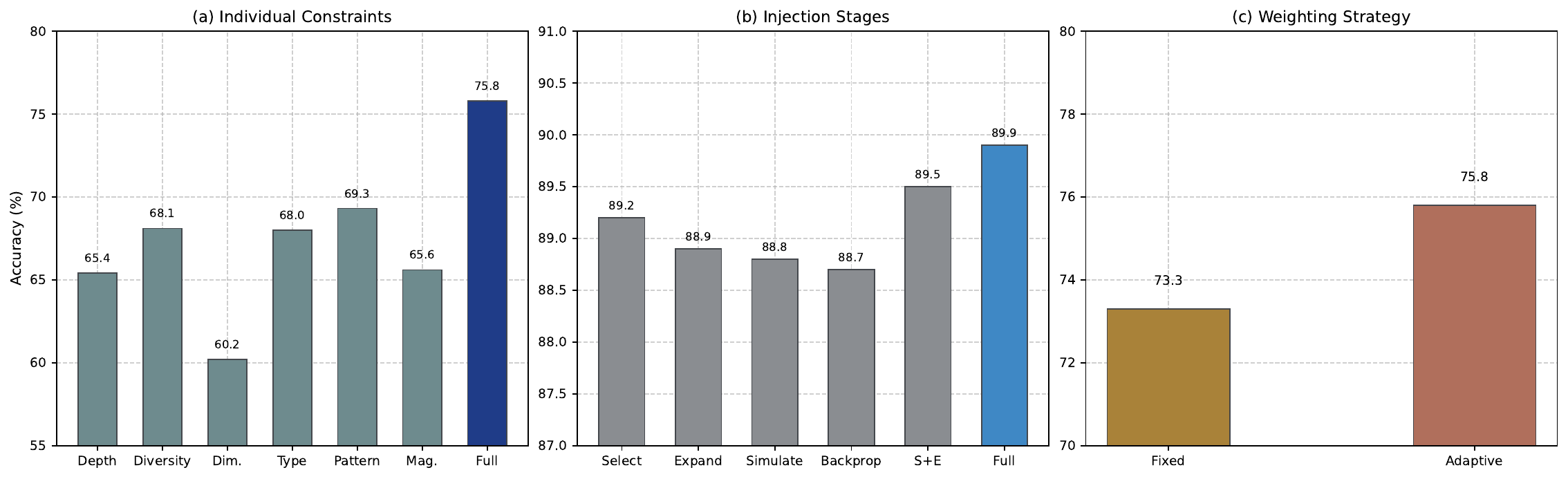}
\caption{Ablation studies on the MATH dataset. (a) Individual constraints: pattern similarity and type compatibility provide strong individual guidance, while dimensional consistency acts as a necessary but restrictive regularizer. (b) Injection stages: full integration across all four stages provides the highest accuracy. (c) Weighting strategy: dynamically updating constraint importance based on performance correlation yields more robust search priors.}
\label{fig:ablation_combined}
\end{figure*}

Our ablation studies on the MATH dataset reveal how individual constraints and their placement contribute to reasoning performance. As shown in Figure~\ref{fig:ablation_combined} and Table~\ref{tab:ablation_constraints}, pattern similarity and type compatibility provide strong individual guidance (69.3\% and 68.0\%), while dimensional consistency acts as a necessary but restrictive regularizer (60.2\%). Full integration achieved 75.8\%, demonstrating that combining constraints improves overall performance. Using GPT-5.2, selection-only reweighting achieved 89.2\%, and full integration across all four stages reached 89.9\% (Table~\ref{tab:ablation_stages}). Adaptive weighting provided a measurable gain over fixed weights (75.8\% vs. 73.3\%, Table~\ref{tab:ablation_adaptive}).

\paragraph{Full ablation results.}
Table~\ref{tab:ablation_constraints} reports the per-constraint ablation results on MATH, and Tables~\ref{tab:ablation_stages}--\ref{tab:ablation_adaptive} report the stage-wise and weighting ablations, including accuracy, constraint satisfaction scores, and computational overhead.

\begin{table}[ht]
\centering
\caption{Numerical results for per-constraint ablation experiments on MATH. All variants use GPT-5-mini. Accuracies and token counts reflect stable averages ($\pm \sigma$) over three runs ($p < 0.05$). Tokens denote average total end-to-end tokens per problem (optimizer+executor) computed from logs.}
\label{tab:ablation_constraints}
\resizebox{\columnwidth}{!}{%
\begin{tabular}{lcccc}
\toprule
\textbf{Constraint Family} & \textbf{Acc. (\%)} & \textbf{Score} & \textbf{Tokens} & \textbf{Cost (\$)} \\
\midrule
Depth Only & 65.4 $\pm$ 1.2 & 0.654 & 35.2k & 0.06427 \\
Diversity Only & 68.1 $\pm$ 0.8 & 0.681 & 34.5k & 0.06299 \\
Dimensional Only & 60.2 $\pm$ 1.5 & 0.602 & 36.4k & 0.06646 \\
Type Only & 68.0 $\pm$ 0.9 & 0.680 & 34.1k & 0.06226 \\
Pattern Only & 69.3 $\pm$ 1.1 & 0.693 & 33.8k & 0.06171 \\
Magnitude Only & 65.6 $\pm$ 1.3 & 0.656 & 34.8k & 0.06354 \\
\midrule
\textbf{All (SCULPT)} & \textbf{75.8 $\pm$ 0.4} & \textbf{0.758} & \textbf{22.5k} & \textbf{0.04108} \\
\bottomrule
\end{tabular}
}
\end{table}

\begin{table}[ht]
\centering
\caption{Detailed numerical results for injection stage ablation experiments on the MATH. All variants use GPT-5.2. Accuracies and token counts reflect stable averages ($\pm \sigma$) over three runs ($p < 0.05$). Tokens denote average total end-to-end tokens per problem (optimizer+executor) computed from logs.}
\label{tab:ablation_stages}
\resizebox{\columnwidth}{!}{%
\begin{tabular}{lcccc}
\toprule
\textbf{Injection Stage(s)} & \textbf{Acc. (\%)} & \textbf{Score} & \textbf{Tokens} & \textbf{Cost (\$)} \\
\midrule
Selection Only & 89.2 $\pm$ 0.4 & 0.892 & 32.5k & 0.29766 \\
Expansion Only & 88.9 $\pm$ 0.5 & 0.889 & 33.8k & 0.30957 \\
Simulation Only & 88.8 $\pm$ 0.6 & 0.888 & 35.2k & 0.32239 \\
Backprop Only & 88.7 $\pm$ 0.5 & 0.887 & 36.5k & 0.33430 \\
Selection + Expansion & 89.5 $\pm$ 0.3 & 0.895 & 28.4k & 0.26011 \\
\midrule
\textbf{SCULPT (All Stages)} & \textbf{89.9 $\pm$ 0.3} & \textbf{0.899} & \textbf{24.1k} & \textbf{0.22073} \\
\bottomrule
\end{tabular}
}
\end{table}

\begin{table}[ht]
\centering
\caption{Comparison of adaptive weighting vs. fixed weights on the MATH validation split. All variants use GPT-5-mini. Accuracies and token counts reflect stable averages ($\pm \sigma$) over three runs ($p < 0.05$). Tokens denote average total end-to-end tokens per problem (optimizer+executor) computed from logs.}
\label{tab:ablation_adaptive}
\resizebox{\columnwidth}{!}{%
\begin{tabular}{lcccc}
\toprule
\textbf{Weighting Strategy} & \textbf{Acc. (\%)} & \textbf{Score} & \textbf{Tokens} & \textbf{Cost (\$)} \\
\midrule
Fixed Weights & 73.3 $\pm$ 1.1 & 0.733 & 25.8k & 0.04711 \\
\midrule
\textbf{Adaptive Weights} & \textbf{75.8 $\pm$ 0.4} & \textbf{0.758} & \textbf{22.5k} & \textbf{0.04108} \\
\bottomrule
\end{tabular}
}
\end{table}

\begin{table}[t]
\caption{Cohort A (GPT-4o-mini executor): accuracy (\%) and efficiency. Optimizer is Claude-3.5-sonnet; executor is GPT-4o-mini-0718. Results for prior methods are taken from~\citet{zhang2024aflow} under matched configurations; variances for those methods are not reported.}
\resizebox{\columnwidth}{!}{%
\begin{tabular}{lccc}
\toprule
\textbf{Method} & \textbf{GSM8K} & \textbf{MATH} \\
\midrule
IO (GPT-4o-mini) & 92.7 & 48.6 \\
CoT~\citep{wei2022chain} & 92.4 & 48.8 \\
CoT SC~\citep{wang2022self} & 92.7 & 50.4 \\
MedPrompt~\citep{nori2023can} & 90.0 & 50.0 \\
MultiPersona~\citep{wang2024unleashing} & 92.8 & 50.8 \\
Self-Refine~\citep{madaan2023self} & 89.6 & 46.1 \\
ADAS~\citep{hu2024automated} & 90.8 & 35.4 \\
AFlow~\citep{zhang2024aflow} & 93.5 & 56.2\\
\textbf{Ours (SCULPT)} & \textbf{93.8 $\pm$ 0.2} & \textbf{57.2 $\pm$ 0.5} \\
\bottomrule
\end{tabular}
}
\label{tab:exec_gpt4omini}
\end{table}

\subsection{Main Results across Executors}
\label{sec:main_results}
SCULPT is evaluated across three distinct executor cohorts to assess generalization and performance on frontier reasoning models. To assess efficiency, average end-to-end token counts per problem (Tokens; optimizer+executor, computed from logs) are reported alongside accuracy.

\paragraph{Cohort A (GPT-4o-mini executor).} 
Using Claude-3.5-sonnet as the optimizer and GPT-4o-mini as the executor, SCULPT achieves strong results as shown in Table~\ref{tab:exec_gpt4omini}. It reaches 93.8\% on GSM8K and 57.2\% on MATH, outperforming automated optimizers like AFlow and ADAS while using fewer end-to-end tokens per problem.

\paragraph{Cohort B (GPT-5-mini executor).} 
Table~\ref{tab:exec_gpt5mini} reports the results for GPT-5-mini. SCULPT improves accuracy on all datasets, including GSM-Hard (v2 split), relative to AFlow (average +1.43\%; MATH: +1.1\%, GSM8K: +2.0\%, GSM-Hard: +1.2\%). The gains are consistent with pruning ineffective exploration paths.

\begin{table}[t]
\centering
\caption{Cohort B (GPT-5-mini executor): accuracy (\%) and efficiency. Results are reproduced using available open-source implementations under the same evaluation protocol.}
\resizebox{\columnwidth}{!}{%
\begin{tabular}{lccccc}
\toprule
\textbf{Method} & \textbf{MATH} & \textbf{GSM8K} & \textbf{GSM-Hard} & \textbf{Avg.} & \textbf{Tokens} \\
\midrule
IO & 68.9 $\pm$ 0.7 & 93.8 $\pm$ 0.4 & 76.7 $\pm$ 0.8 & 79.8 & 915 \\
Self-Refine~\citep{madaan2023self} & 65.4 $\pm$ 0.9 & 91.7 $\pm$ 0.6 & 73.8 $\pm$ 1.1 & 77.0 & 3,240 \\
MultiPersona~\citep{wang2024unleashing} & 71.5 $\pm$ 0.8 & 97.4 $\pm$ 0.3 & 80.6 $\pm$ 0.7 & 83.2 & 7,850 \\
ADAS~\citep{hu2024automated} & 60.2 $\pm$ 1.1 & 92.9 $\pm$ 0.6 & 65.8 $\pm$ 1.3 & 73.0 & 45,038 \\
AFlow~\citep{zhang2024aflow} & 74.7 $\pm$ 0.8 & 95.6 $\pm$ 0.5 & 83.2 $\pm$ 0.9 & 84.5 & 30,025 \\
\textbf{Ours (SCULPT)} & \textbf{75.8 $\pm$ 0.4} & \textbf{97.6 $\pm$ 0.2} & \textbf{84.4 $\pm$ 0.5} & \textbf{85.9} & \textbf{22,519} \\
\bottomrule
\end{tabular}
}
\label{tab:exec_gpt5mini}
\end{table}

\paragraph{Cohort C (GPT-5.2 executor).} 
To evaluate the scalability of SCULPT on frontier reasoning models, we conduct experiments using GPT-5.2 as the core executor. All compared methods use GPT-5.2 as the executor with identical experimental settings (data splits, random seeds, evaluation procedures). As shown in Table~\ref{tab:exec_gpt52}, SCULPT performs strongly, particularly in complex domains like MATH and GSM-Hard where it improves over AFlow. For GSM8K, accuracy across high-performing methods is close to a ceiling (97.8\% vs 98.0\%), so marginal differences can be within stochastic variance. Meanwhile, pruning an average of 34.2\% of branches yields meaningful efficiency gains even when accuracy differences are small.

\begin{table}[t]
\centering
\caption{Cohort C (GPT-5.2 executor): accuracy (\%) and efficiency.}
\resizebox{\columnwidth}{!}{%
\begin{tabular}{lccccc}
\toprule
\textbf{Method} & \textbf{MATH} & \textbf{GSM8K} & \textbf{GSM-Hard} & \textbf{Avg.} & \textbf{Tokens} \\
\midrule
IO & 81.8 $\pm$ 0.5 & 96.3 $\pm$ 0.3 & 80.2 $\pm$ 0.6 & 86.1 & 1,120 \\
Self-Refine~\citep{madaan2023self} & 80.5 $\pm$ 0.6 & 97.0 $\pm$ 0.2 & 79.9 $\pm$ 0.8 & 85.8 & 4,150 \\
MultiPersona~\citep{wang2024unleashing} & 75.3 $\pm$ 1.2 & 96.8 $\pm$ 0.4 & 84.0 $\pm$ 0.6 & 85.4 & 11,240 \\
ADAS~\citep{hu2024automated} & 78.8 $\pm$ 0.9 & 92.5 $\pm$ 0.5 & 70.2 $\pm$ 1.1 & 80.5 & 48,284 \\
AFlow~\citep{zhang2024aflow} & 88.7 $\pm$ 0.6 & 98.0 $\pm$ 0.1 & 87.0 $\pm$ 0.7 & 91.2 & 32,189 \\
\textbf{Ours (SCULPT)} & \textbf{89.9 $\pm$ 0.3} & \textbf{97.8 $\pm$ 0.2} & \textbf{88.3 $\pm$ 0.4} & \textbf{92.0} & \textbf{24,142} \\
\bottomrule
\end{tabular}
}
\label{tab:exec_gpt52}
\end{table}

\subsection{Reasoning Stability and Efficiency}
\paragraph{Search Stability and Efficiency.} 
SCULPT improves search stability and computational efficiency in our experiments. The pruning rate of \textbf{34.2\%} (averaged across 15 rounds and three datasets, std: 3.1\%) filters candidates below the depth-aware threshold $\tau(d)$, reducing token usage. In our measured setup, SCULPT achieves \textbf{28\% lower optimization time} and \textbf{31\% lower API expenditure} through search space regularization. Profiling indicates that the symbolic checks add negligible overhead (<0.1\% of total wall-clock time). SCULPT also exhibits improved stability: validation score standard deviation is 58\% of AFlow's ($\sigma_{\text{SCULPT}} / \sigma_{\text{AFlow}} = 0.58$), corresponding to a 66\% variance reduction.

\paragraph{Qualitative Analysis.} 
On a sample MATH problem involving unit conversions, AFlow fails by adding magnitudes with inconsistent units, while SCULPT identifies the dimension mismatch early via the Instruction Hub. The aggregate compliance $C_{\text{total}}$ drops for the erroneous branch, triggering early pruning and steering MCTS toward a valid conversion routine, confirming SCULPT's role as a domain-aware regularizer.

\FloatBarrier
\section{Conclusion}
In this paper, we introduce SCULPT, a constraint-guided MCTS framework that transitions agentic workflow optimization from unguided stochastic sampling to ordered, domain-aware exploration. By systematically integrating domain-aware constraints into the selection, expansion, and backpropagation phases, SCULPT effectively regularizes the search space of complex mathematical reasoning. Empirical evaluations demonstrate that SCULPT transcends traditional performance gains, achieving a \textbf{34.2\%} reduction in implausible branch expansion and a \textbf{31\%} decrease in cumulative API expenditure. Furthermore, the integration of compliance-aware rewards yields a \textbf{66\%} reduction in search variance, significantly enhancing the robustness of automated workflow synthesis. These results underscore the potential of domain-aware constraint integration in optimizing long-horizon reasoning. Future work will investigate the portability of SCULPT to other high-precision domains, such as formal verification and legal logic, where structured constraints are essential for efficient pathfinding.

\section*{Limitations}

Our study focuses on mathematical reasoning; generalization requires task-specific constraint design. Symbolic checks may miss deeper semantic relations, especially in geometry. The structural pattern component is refined during the optimization phase, which may limit immediate applicability to problem distributions that differ substantially from the validation set. Magnitude constraints require execution traces and cannot contribute to early pruning. More principled adaptation and profiling are left for future work.

\section*{Acknowledgements}
This work was supported by the Natural Science Foundation of Jilin Provincial Science and Technology Department, China, under Grant YDZJ202201ZYTS553.

\bibliography{custom}

\appendix

\section{Reproducibility and Implementation Details}
\label{sec:appendix_repro}

\paragraph{Configuration summary.}
The random seed is fixed to \textbf{42}. For each dataset, the same validation/test split as the corresponding baseline runs is used, and the optimizer--executor pair is kept fixed within each cohort. Hyperparameters follow Section~\ref{headings}: \(\epsilon=0.01\), \(\tau(d)=\max(\tau_{\min},\tau_0-kd)\) with \(\tau_0=0.6\), \(\tau_{\min}=0.3\), \(k=0.05\), and selection shaping strength \(\lambda=0.5\). These values are intentionally chosen as simple, conservative defaults to stabilize search and keep the method lightweight (avoiding extensive per-dataset tuning or additional control knobs); more exhaustive sensitivity sweeps are left for future work. Unless otherwise noted (e.g., ablations), constraint weights are initialized uniformly and then updated by Eq.~\eqref{eq:adaptive_weighting} after a 5-round warm-up.

\paragraph{Workflow representation for constraint scoring.}
Each workflow instance is represented as a structured program graph with a lightweight view exposed to SCULPT as a dictionary-like state. The state contains: (i) depth \(d(w)\), (ii) an operator multiset/histogram (used by \(C_{\mathcal{V}}\) and \(C_{\mathcal{P}}\)), and (iii) when available from execution traces, a summary of intermediate magnitudes \(\max(|\mathcal{X}_w|)\) (used by \(C_{\mathcal{M}}\)). Static constraints are computed without executing the workflow; execution-aware constraints are computed from rollout traces.

\paragraph{Constraint scoring details.}
Per-family scores \(C_i\in[0,1]\) are computed and aggregated using Eq.~\eqref{eq:compliance_agg}.
\begin{itemize}
  \item \textbf{Depth control \(C_{\mathcal{D}}\).} We penalize workflows exceeding a depth threshold \(d_{\max}\) using Eq.~\eqref{eq:depth_penalty}.
  \item \textbf{Diversity \(C_{\mathcal{V}}\).} We compute the normalized Shannon entropy over operator usage as in Eq.~\eqref{eq:diversity_reward}.
  \item \textbf{Dimensional consistency \(C_{\mathcal{U}}\).} We apply a lightweight compatibility check when the workflow state exposes unit-like tags; if unit metadata is unavailable, this component defaults to a neutral score so that missing tags do not by themselves trigger early pruning.
  \item \textbf{Type compatibility \(C_{\mathcal{T}}\).} We check local domain/type constraints for common mathematical operators (e.g., \(\sqrt{\cdot}\) inputs non-negative; \(\log(\cdot)\) inputs positive; and consistent scalar/vector/matrix usage when shapes are exposed).
  \item \textbf{Pattern similarity \(C_{\mathcal{P}}\).} We compute similarity between the workflow operator signature and a motif library for the detected problem class; when no motif matches, we use a neutral prior (0.5) in early rounds.
  \item \textbf{Magnitude sanity \(C_{\mathcal{M}}\).} We downweight workflows that produce numerically implausible intermediate values according to Eq.~\eqref{eq:magnitude_sanity}; this score is computed only when execution traces provide \(\mathcal{X}_w\).
\end{itemize}

\paragraph{SCULPT-guided MCTS (pseudocode).}
The pseudocode below summarizes the four-stage loop described in Section~\ref{headings}, including depth-aware pruning and compliance-weighted selection/backpropagation.

\noindent\begin{minipage}{\columnwidth}
\scriptsize\ttfamily\raggedright
Inputs: root workflow w0, budget T, depth threshold schedule tau(d); constraint families \{Ci\}; weights \{wi\}.\\
Selection: select node v by maximizing (Q + c*U)*exp(lambda*C\_total(v)).\\
Expansion: generate edits \{wk\} from v; keep wk with C\_total(wk) >= tau(depth(v)); if empty, keep argmax\_k C\_total(wk).\\
Simulation: run wk to get reward R and traces X\_wk; compute C\_M(X\_wk); update C\_total(wk).\\
Backprop: backpropagate R*C\_total(wk) through the path.\\
Adapt: update weights wi using corr(Ci, validation reward) (Eq. 7).\\
Return: best workflow under budget.
\end{minipage}

\paragraph{Examples of thresholded expansion.}
During expansion, SCULPT filters candidate workflow edits using the depth-aware gate \(\tau(d)\) in Eq.~\eqref{eq:tau_threshold}. Table~\ref{tab:prune_examples} lists representative candidate types and the corresponding gate outcome; these examples illustrate the criterion \(C_{\text{total}} \ge \tau(d)\) and the dominant constraint families responsible for rejection.

\begin{table}[t]
\centering
\caption{Representative examples of expansion candidates filtered by the depth-aware threshold.}
\label{tab:prune_examples}
\setlength{\tabcolsep}{3pt}
\small
\resizebox{\columnwidth}{!}{%
\begin{tabular}{p{0.62\linewidth}cc}
\toprule
\textbf{Candidate edit (summary)} & \textbf{Gate} & \textbf{Dominant reason(s)} \\
\midrule
Increase search depth by adding redundant self-verification loops without new operators & pruned & \(C_{\mathcal{D}}\downarrow\), \(C_{\mathcal{V}}\downarrow\) \\
Introduce an operation that violates a local domain rule (e.g., applying \(\log(\cdot)\) to a non-positive intermediate) & pruned & \(C_{\mathcal{T}}\downarrow\) \\
Add an aggressive numerical transformation that produces implausible intermediate magnitudes during rollout & pruned & \(C_{\mathcal{M}}\downarrow\) \\
\bottomrule
\end{tabular}
}
\end{table}

\section{Additional Analysis: Error Categories and Representative Cases}
\label{sec:appendix_error}

\paragraph{Setup.}
SCULPT is compared against an MCTS baseline that disables the proposed constraint-guided pruning on a 100-problem MATH evaluation subset (a fixed slice used for lightweight analysis). Problems are grouped into coarse error categories using keyword templates (e.g., geometry/angles/areas; counting/probability), and the number of baseline errors eliminated within each category is summarized.

\begin{table}[t]
\centering
\caption{Error-category comparison on a 100-problem MATH subset (same questions for the MCTS baseline and SCULPT). ``Reduced'' counts the number of baseline errors eliminated by SCULPT within each category; categories are coarse templates and therefore do not cover all problems.}
\label{tab:error_category}
\setlength{\tabcolsep}{3pt}
\small
\resizebox{\columnwidth}{!}{%
\begin{tabular}{lrrrr}
\toprule
\textbf{Category} & \textbf{\#} & \textbf{Baseline Wrong} & \textbf{SCULPT Wrong} & \textbf{Reduced} \\
\midrule
Counting/Probability & 26 & 12 & 9 & 3 \\
Geometry/Angle/Area & 12 & 10 & 8 & 2 \\
Algebra/Equations & 7 & 2 & 2 & 0 \\
\midrule
\textbf{Total (overall)} & 100 & 49 & 44 & 5 \\
\bottomrule
\end{tabular}
}
\end{table}

\paragraph{Representative corrected cases.}
In this subset, reductions are concentrated in counting/probability and geometry (Table~\ref{tab:error_category}). Common corrected patterns include enforcing complete case enumeration (avoiding missed cases/double counting) and geometry conventions (e.g., consistent use of standard triangle ratios and non-negative area expressions).

\end{document}